\documentclass[a4paper,twoside]{article}

\usepackage{epsfig}
\usepackage{subcaption}
\usepackage{calc}
\usepackage{amssymb}
\usepackage{amstext}
\usepackage{amsmath}
\usepackage{amsthm}
\usepackage{multicol}
\usepackage{pslatex}
\usepackage{apalike}
\usepackage{algorithm2e}
\usepackage[bottom]{footmisc}
\usepackage{SCITEPRESS}     
\usepackage{cleveref}
\usepackage{booktabs}
\usepackage{stfloats}
\usepackage[font=small,skip=4pt]{caption}

\usepackage{microtype}
\begin{document}

\title{Geometric 2D Scene Graph Generation}

\if 0
\author{
\authorname{Anonymous Authors}
\affiliation{Paper under double-blind review}
%\email{anonymous@example.com}
\email{\mbox{}}
}
\fi

\author{\authorname{Christoph Jahn\sup{1,2}\orcidAuthor{0009-0004-1511-9191}, Urs Waldmann\sup{2}\orcidAuthor{0000-0002-1626-9253} and Bastian Goldl\"ucke\sup{2}\orcidAuthor{0000-0003-3427-4029}}
\affiliation{\sup{1}Mercedes-Benz AG, Sindelfingen, Germany}
\affiliation{\sup{2}Department of Computer and Information Science, University of Konstanz, Konstanz, Germany}
\email{christoph.jahn@mercedes-benz.com, \{urs.waldmann, bastian.goldluecke\}@uni-konstanz.de}
}


\abstract{In production processes for consumer products,
assembly instructions are essential not only for planning but also for executing the production process.
Likewise in robotics, it is crucial for an assembly robot to understand how components fit together and can be assembled.
To facilitate these tasks, we contribute a method for constructing scene graphs to represent and characterize 
assembly relationships between components.
Our approach does not rely on semantic data and is capable of handling a very small dataset.
To realize this, the output of a Faster R-CNN model is used to create geometric representations, which are then processed by a transformer architecture to generate an adjacency matrix. This matrix serves as input to a Siamese network that uses message passing based on an attentional graph convolutional network (aGCN) architecture to characterize the connections between the components.
We validate our method on a study dataset of toy model components which can be assembled into transportation vehicles.}

\onecolumn
\maketitle
\normalsize
\setcounter{footnote}{0}

\section{\uppercase{Introduction}}
\label{sec:introduction}

\begin{figure*}[!ht]
  \centering
   {\epsfig{file = 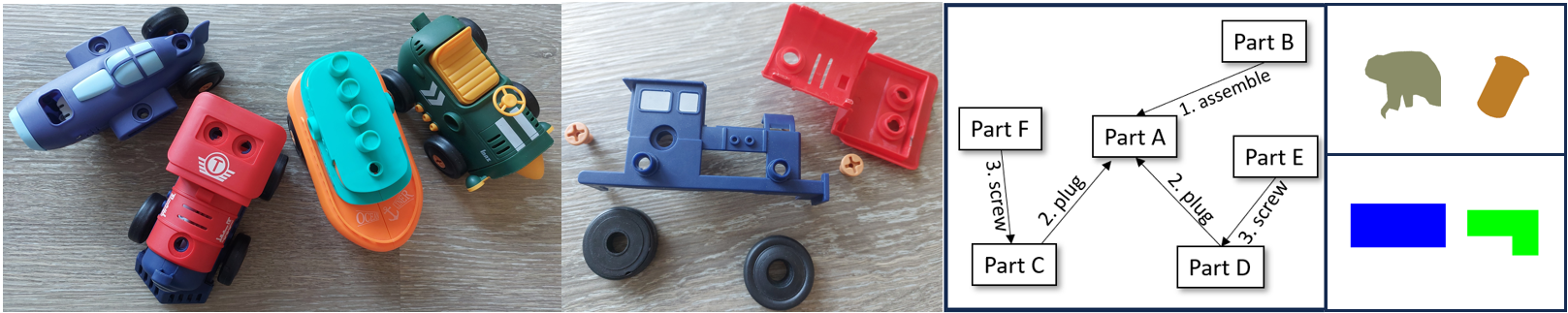, width = 15.8cm}}
    \vspace{1mm}
    \begin{minipage}{\textwidth}
        \raggedright
        \fontsize{8pt}{9pt}\selectfont
        \hspace{0.7cm} (a) Vehicles (Plane, Train, Boat, Car)     
        \hspace{1.1cm} (b) Model Input: Parts     
        \hspace{1.2cm} (c) Model Output: Graph
        \hspace{0.9cm} (d) Primitives
    \end{minipage}
    \caption{To evaluate our approach, we conduct a case study (Fig.~\ref{fig:1}a–d) where the model is trained on components from three toy vehicles (train, boat, plane) and must construct a geometric scene graph for an unseen vehicle (car). The graph encodes both the assembly structure and sequence. Training data is augmented with simple geometric primitives to improve generalization (Fig.~\ref{fig:1}d).}
    \label{fig:1}
\end{figure*}

Scene graph generation is a widely used technique for visual scene understanding, allowing scenes in both 2D images and 3D data to be described through objects and the relationships between them \cite{johnson2015image,sadeghi2011recognition,lu2016visual,wald2020learning,dhamo2021graph,Sarkar_2023_ICCV}. Objects are represented as nodes, while their relationships are depicted as edges, typically expressed in the form of subject–predicate–object triplets. This representation improves contextual reasoning by modeling spatial and functional object constellations.
In the context of assembly modeling, a distinction can be made between generating scene graphs for assembled and disassembled configurations. This work focuses on disassembled parts and presents a method for transforming such components (see Fig.~\ref{fig:1}b) into a scene graph representation (see Fig.~\ref{fig:1}c). This provides a solution to the practical problem of determining which components belong together and how they are connected.

End-to-end approaches to scene graph prediction~\cite{cong2023reltr}, which simultaneously detect objects, identify their relationships and predict relationship triplets directly from an image, require a substantial amount of training data and a carefully annotated triplet-labeled dataset, which is not feasible in our envisioned scenario.
As a remedy, previous works adopt a two-step pipeline, where in a first step objects or region proposals are detected
with pre-trained networks before further processing~\cite{li2017scene,kraehenbuehl2017scene,yang2018graph,johnson2015image,sadeghi2011recognition,lu2016visual,zellers2018neural}.
Our approach is closely aligned with the two-step method proposed by Jianwei Yang et al.~\cite{yang2018graph}, where an adjacency matrix is predicted as an intermediate representation.
This matrix encodes connectivity of all pairs of components and serves as backbone of the resulting graph.

However, Yang et al.~\cite{yang2018graph} rely on semantic priors to construct the adjacency matrix.
While semantic cues often provide valuable
contextual relationships and structural hints that support graph inference~\cite{yang2018graph,cong2023reltr},
they are frequently unavailable or uninformative in practical applications such as furniture or vehicle assembly.
We therefore expand upon~\cite{yang2018graph} and 
propose a three-step prediction process that explicitly separates the structural prediction of the adjacency matrix from the semantic tasks of predicting component labels, permutations, and assembly sequences in a transformer-based architecture.
This separation allows the model to first infer structural relationships independently, before reasoning about semantic or procedural aspects. 

\paragraph{Contributions.}
We address the challenge of scene graph prediction in component assemblies under small dataset conditions and without semantic information. 
To overcome the limitations of conventional one- and two-step methods, we propose a three-step architecture that separates structural inference from semantic and procedural reasoning.
Our transformer-based model predicts adjacency matrices purely from geometric cues, enabling robust graph construction with minimal labeled data even when semantic cues are missing or sparse.
Permutations ensure that each component is used only once, while sequences determine the correct assembly order. All label predictions are conditioned on the same adjacency matrix, ensuring consistent modeling of binary relations.
To further encourage robust generalization to unknown configurations, we augment the training data with primitive shapes derived from color and geometry to emphasize relevant geometric features~\cite{Gong_2021_CVPR}.
In addition, both the transformer architecture and the loss function are adapted to more effectively capture and supervise the structural relationships between geometric objects within the scene graph.
We create a custom dataset of toy vehicle assemblies and validate our approach on a case study involving unseen vehicle types,
see~\cref{fig:1}.

\section{\uppercase{Related Work}}
\label{sec:related-work}

In recent years, several approaches have been developed to address the task of scene graph generation. Traditional methods employ backbones such as Mask R-CNN, Faster R-CNN or DETR~\cite{yang2018graph,cong2023reltr} to generate the objects for the actual scene graph generation network. These networks are typically based on recurrent neural networks (RNNs)~\cite{kraehenbuehl2017scene}, LSTMs~\cite{zellers2018neural}, or more recently, transformer architectures~\cite{cong2023reltr,li2023sgtr,sortino2023transformer,SGTR+2023}, message passing mechanisms and attention models~\cite{yang2018graph,Yang2018AttentiveRN} enabling the integration of contextual information. Such context helps components interpret their surroundings and supports the recognition of graph structures, which is crucial for improving prediction accuracy.

Two-step scene graph models first detect objects and then infer graph structures~\cite{kraehenbuehl2017scene,yang2018graph}. In contrast, one-step methods jointly predict object classes, bounding boxes, and relationships, integrating detection into a single pipeline~\cite{cong2023reltr}. These models often treat the task as set prediction, allowing unordered subject–relation–object triplets without assuming a fixed output and avoiding explicit prediction of non-existent relations.

Unlike set prediction models like RelTR~\cite{cong2023reltr}, our architecture uses adjacency matrices instead of predefined subject–object pairs. This has two main advantages. First, set prediction typically requires large, fully annotated datasets. In contrast, our method evaluates all possible pairwise connections, ensuring coverage without dense annotations. Although predicting scene graphs via set prediction is feasible, it would require much more data compared to adjacency matrix prediction with low-dimensional features \cite{hoang2023graph,rezatofighi2017deepsetnet,rezatofighi2020learn}. Second, set prediction may miss rare relations or waste output slots on false positives. Due to our limited dataset, we adopt a three-stage pipeline (see Fig.~\ref{fig:2}) to reduce model complexity and enable subtask specialization. Semantic and visual information is typically processed via MLP backends to predict and characterize object relationships~\cite{cong2023reltr}.

Semantic knowledge is widely used across models \cite{yang2018graph,cong2023reltr,lu2016visual}. The influence of semantic input varies: it can play a central role \cite{yang2018graph} or be less prominent \cite{cong2023reltr}. 
While it is helpful for label prediction, too much can overshadow visual features. 

Handling graph structures poses a challenge, as models must distinguish between bipartite and unidirectional graphs. We use unidirectional graphs for subject–object relations, with adjacency matrix rows as subjects and columns as objects. This avoids ambiguity in role assignment, which bidirectional graphs can cause. To address imbalanced node and edge frequencies, focal loss and sampling strategies are commonly used~\cite{Lin_2020_CVPR,Tang2020UnbiasedSGG}.

Scene graph generation methods besides GCNs \cite{xu2020understanding} and transformers \cite{cong2023reltr,Zhu2020DeformableDT,li2021bipartite} also include diffusion models \cite{liu2024joint}. However, diffusion models are primarily designed for continuous, generative tasks and typically require large-scale datasets and extensive training resources. Our method employs a transformer, whose self-attention mechanism inherently reflects an adjacency-matrix-like structure. Unlike GCN methods with predefined graphs, we first predict the adjacency matrix before applying a GCN. Scene graphs are applied in domains like autonomous driving \cite{zhang2024graphad}, robotics \cite{GRID2023} and visual question answering \cite{Lin_2020_CVPR,tang2019learning}.

\section{\uppercase{Geometric graph generation model}}
\label{sec:Approach}
\begin{figure*}[!ht]
  \centering
   {\epsfig{file = 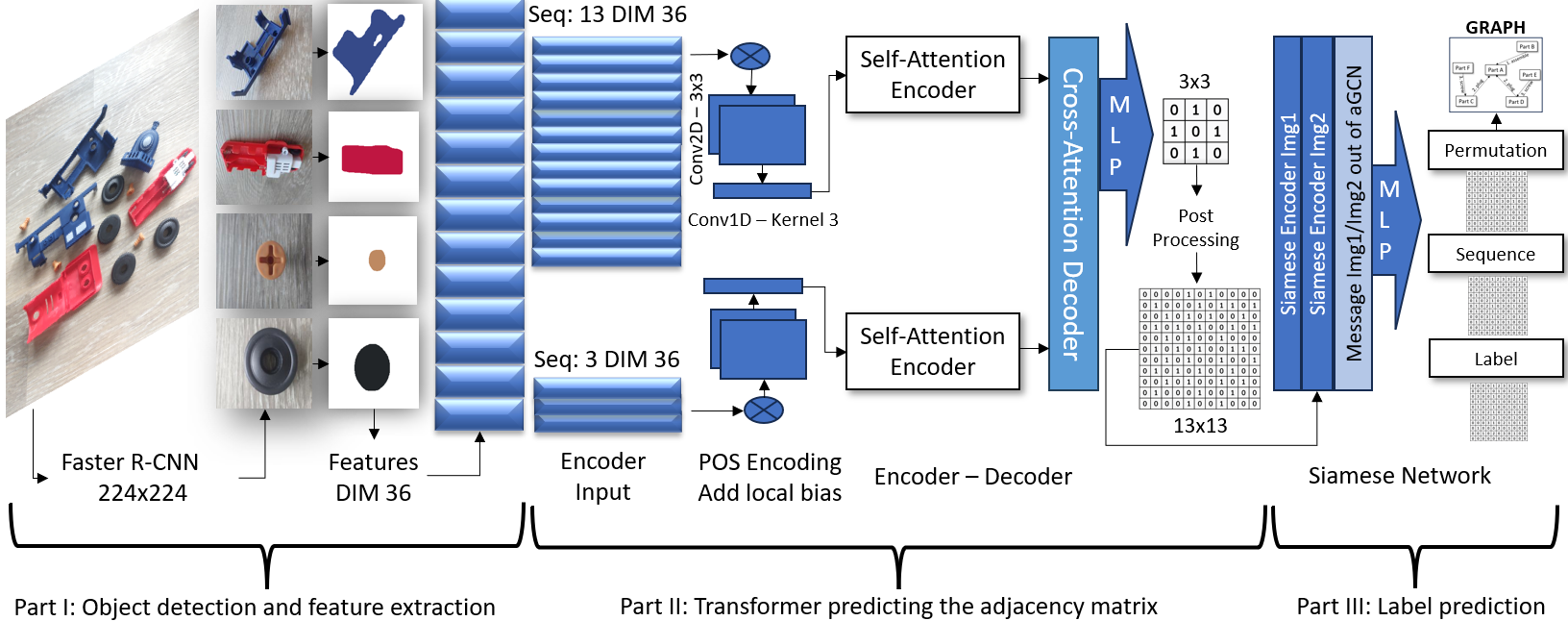, width = 15.8cm}}
    \caption{Our three-step pipeline: First, component bounding boxes and geometric features are extracted without semantic labels; second, a transformer predicts the binary relations based on geometric relations; third, a GCN-based Siamese network infers relation labels, permutations and assembly sequences of each connected pair in the adjacency matrix.}
    \label{fig:2}
\end{figure*}

Overall, our model outlined in Fig.~\ref{fig:2} follows a three-step pipeline, with each step representing a core architectural component. The input of an image with disassembled components is processed over three steps to the output of a corresponding geometric scene graph. 
First, a Faster R-CNN \cite{yang2018graph} detects component bounding boxes and geometric features get extracted. As object classes are not identified, the approach remains robust to unknown parts by avoiding semantic dependencies. Second, a transformer with cross-attention predicts binary relations based on geometric relationships. Object–subject labels are not used at this stage, allowing structural reasoning to rely solely on geometry. Third, a Siamese network with GCN-based message passing predicts relation labels, permutations and assembly sequences for all components.
The predicted relations in final the scene graph can be interpreted as an assembly instruction (see Fig.~\ref{fig:1}c). 

\subsection{Object detection and feature extraction}
The input of this first step of the model is an image of the disassembled components.
A Faster R-CNN extracts an individual image for each component, which is processed by a pretrained Mask R-CNN 
to extract its colored shape.
The pre-processing increases robustness by minimizing background, shading and feature detail artifacts~\cite{ilyas2019adversarial,sehwag2020background}.
It also allows augmentations with geometric primitives (see Fig.~\ref{fig:1}d),
which broaden the data distribution more effectively than direct graph augmentations.
The resulting shape images are passed to a pretrained Vision Transformer (ViT-B/32) \cite{dosovitskiy2021an} to generate a feature vector
with feature dimension~$F$.

\subsection{Transformer predicting the adjacency matrix}
The following section details the transformer-based graph construction (see Fig.~\ref{fig:2} - Part II). 
The high amount of connections and the imbalance of negative pairs present a modeling challenge. To address this, we split the graph into smaller subgraphs and later merge partial predictions~\cite{mao2022graph,wang2021patchgnn,correia2023sparse}.
We reduce the amount of components used in the prediction of a subgraph with an encoder–decoder transformer.
Cross-attention with all given components ensures each subgraph prediction maintains global context (see Fig.~\ref{fig:2}).

Our contextual strategy aims to improve accuracy by incorporating neighbor and global graph information. Since transformers emphasize global context and may underutilize local spatial cues~\cite{raghu2021do}, we apply depthwise convolutions along feature and sequence dimensions before transformer encoding~\cite{kim2024depthwise} to enhance the input with localized spatial features. For this, the positional encodings are reshaped to match the expected convolutional input format. 

Two encoders are employed to generate the input embeddings for the decoder using self-attention, where queries ($Q$), keys ($K$) and values ($V$) are derived from the input representations.
For all components of the vehicle graph \( Q_{\text{all}},\,K_{\text{all}},\,V_{\text{all}} \in \mathbb{R}^{n \times F} \)
and for the reduced components of the vehicle subgraph \(Q_\text{subgraph},\,K_\text{subgraph},\,V_\text{subgraph} \in \mathbb{R}^{s \times F}\)
two self-attention encoder are used, where \(n\) and \(s\) represent the sequences and \(F\) the feature dimension of the transformer.
To enhance graph understanding and capture interactions between all graph features, the cross-attention variable \(e\) is introduced as an edge-augmented attention score within the Edge Graph Transformer (EGT) framework \cite{hussain2021global}, enabling non-linear modeling of edge-to-edge relationships according to
\begin{equation}
\text{CrossAtt}_a(Q_a, K'_a, V'_a) = \sigma \left( \frac{Q_a {K'_a}^T}{\sqrt{d_k}} + e_{i,j} \right) V'_a.
\end{equation}
Here, \( \text{CrossAtt}_a \) denotes the cross-attention operation and \( Q_a \) denotes the encoder embedding of the reduced parts, which serves as the target input for the decoder.  
The matrices \( K'_a \) and \( V'_a \) represent the encoder embeddings of all parts and are used as the source input to the decoder.  
The attention scores \( e_{i,j} = \mathrm{MLP}([q_i \,\|\, k_j]) \) are derived by passing the \( i \)-th row \( q_i \) of \( Q_a \) and the \( j \)-th row with \( k_j \) of \( K'_a \) through an MLP and \( \sigma \) denotes the softmax function.

Inference is run on sub-samples, each representing one subgraph.
In a post-processing step, the resulting partial predictions are merged into a full \( n \times n \)
adjacency matrix. This aggregation captures a certain fraction of relevant connections, given by the percentage of draws
compared to total number of ordered component pairs, which is set as a confidence threshold hyperparameter~$\mathcal{P}$.

\subsection{Characterization of geometric relationships}

The predicted adjacency matrix is passed to a Siamese network with GCN-based message passing to classify component relationships, see~\cref{fig:2}.
This step predicts labels describing assembly actions, permissible permutations and the correct assembly sequence.
A challenge lies in handling permutation constraints, e.g. a wheel may only be assembled with one screw. Encoding such logic directly into the binary adjacency matrix would require the attention mechanism to distinguish valid from invalid permutations.

However, current transformer models struggle with such contextual constraint learning \cite{lee2019set,huang2021permutation}. Therefore, permutations are predicted separately as individual component labels.
All positive connections in the predicted adjacency matrix are extracted and each component pair $x_i$, $x_j$ is processed by the Siamese encoder. The encoder output is combined with the corresponding GCN messages, where all features are propagated using the predicted adjacency matrix \cite{yang2018graph,Chen_2021_CVPR}.

In detail, the output predictions $\hat{y}_{ij}$ are computed according to
\begin{equation}
\begin{aligned}
\hat{y}_{ij} &= \mathrm{MLP}(x_i U \parallel x_j U \parallel V(Z_i + Z_j))\\
\text{where }
Z &= W_S X + \mathrm{softmax}(\mathrm{attn}(h)) \odot h\\
\text{and }
h &= X W_O + A (X W_H).
\end{aligned}
\end{equation}
Above, we denote $W_O$ as the aggregation weight for all objects, $W_H$ as the neighbor aggregation weight and $W_S$ as the skip connection weight. 
The Siamese encoder weight is represented by $U$, while $V$ corresponds to the readout projection weight. 
The adjacency matrix of the graph is given by $A \in \{0,1\}^{n \times n}$. 

In terms of operators and tensor representations, $\mathrm{attn}$ indicates the linear attention projection, $\odot$ denotes element-wise multiplication, and $\parallel$ represents vector concatenation. 
The classifier applied to node embeddings is denoted by $\mathrm{MLP}$, the node features are contained in $X$ and the output of the graph convolutional network is represented by $Z$.

Finally, as a postprocessing step, the predicted permutations are used to filter the label and sequence matrices, retaining only those connections that are consistent with valid assembly relationships between the labels and sequences. This allows the scene graph to be constructed as shown in~\cref{fig:1,fig:2}.

\subsection{Loss functions}
The following section presents the loss function used for the transformer component of the architecture, while the GCN module employs a standard cross-entropy loss. A key challenge is enabling the model to understand graph structures, particularly how graph elements are related to one another based on their attributes and functional context. On small datasets models often overfit by memorizing artifacts or specific patterns.
The core issue is that global node–edge relationships must be inferred from limited input, typically just component features.
To enhance graph reasoning, carefully designed augmentations, geometric cues and architectural biases (e.g., local and edge-aware) are essential.

Class imbalance is addressed using a focal loss~$\mathcal{L}_{\text{focal}}$, which extends cross-entropy to emphasize difficult cases
and helps the model capture complex structures despite skewed label distributions \cite{cong2023reltr}.
It is computed as
\begin{equation}
\begin{aligned}
\mathcal{L}_{\text{focal}} &= y \mathcal{L}_{\text{pos}} + (1 - y) \mathcal{L}_{\text{neg}},\\
\text{where }\mathcal{L}_{\text{pos}} &= - \alpha (1 - p)^{\gamma} \log(p + \varepsilon)\\
\text{and }\mathcal{L}_{\text{neg}} &= - (1 - \alpha) p^{\gamma} \log(1 - p + \varepsilon).
\end{aligned}
\end{equation}
Above, we use the predicted probability \( p \in [0,1] \), the binary ground truth \( y \in \{0,1\} \) for each entry in the adjacency matrix, a weighting factor \( \alpha = 0.55 \) for the positive class, a focusing parameter \( \gamma = 2.0 \), and a small constant \( \epsilon = 10^{-10}\ \) for numerical stability.

To complement the binary focal loss, we apply additional regularization. While contrastive losses are often used to align embeddings of nodes or edges from the same class ~\cite{khosla2020supervised,yang2023local},
our experiments with local GCL and supervised contrastive loss (SCL) showed no improvement.
A key limitation is the reliance of SCL on large batch sizes to distinguish
positive pairs~\cite{you2020graph,liu2022simple,yang2023local,khosla2020supervised}, which is
unsuitable for our small vehicle dataset used in the case study.

Instead, we use classical Laplacian regularization and a group loss. Both are applied to the transformer decoder embeddings prior to the MLP classification layer for the adjacency matrix.
Laplacian regularization encourages smoothness by penalizing large embedding differences between adjacent nodes \cite{Zeng2019DGLR,Belkin2006Laplacian} according to
\begin{equation}
\mathcal{L}_{\text{laplace}} = \frac{1}{B} \sum_{b=1}^B \mathrm{Tr}(H_b^\top L_b H_b),
\end{equation}
where $H_b$ 
is the transformer decoder embedding obtained before the MLP,
and $L_b = D_b - A_b$ is the graph Laplacian computed from the
diagonal degree matrix $D_b$ of the $b$th graph in the current training batch of size $B$,
and the normalized, symmetric adjacency matrix $A_b$, both available during training.

The proposed group loss strengthens graph understanding and supports underrepresented
classes~\cite{Elezi_2020_ECCV}.
For example, in vehicle graphs, common parts like wheels or screws are easy to classify due to clear geometrical structure and frequency of occurrence.
Attachment components however, appear less often and have ambiguous shapes.
Grouping them thus improves semantic reasoning. In this example, if there is a component with low classification
certainty then it is likely one of the attachment components.
The group loss is computed according to
\begin{equation}
\mathcal{L}_{\text{group}} =
\left(
\frac{1}{N} \sum_{x_{j=1}}^{N} \sigma(z_{j}) M_{j}
- 1
\right)^2,
\end{equation}

where $\sigma(z_j)$ denotes the sigmoid-activated logit for the \( j \)-th node $x_{j}$, with $N$ as the set of all nodes in the graph, indicating group membership and $M_j \in {0, 1}$ is a binary mask indicating whether node $x_{j}$ belongs to the group.

Finally, all loss functions are summed with appropriate weights to obtain the final
training objective. Here, in addition to the ones defined above, we also include
a standard cross-entropy loss for multiclass classification
for the Siamese network with GCN message passing.

\section{\uppercase{Evaluation}}
\label{sec:LabelPropagation}

In this section, we evaluate the implementation of our model (see Section~\ref{sec:Approach}) using a dataset of four toy vehicles illustrated in Fig.~\ref{fig:1}. Due to the limited availability of labeled data, this work is based on a small dataset and should be understood as a conceptual study. The architectural parameters are detailed in Fig.~\ref{fig:2}.

\begin{table*}[b]
\caption{Quantitative evaluation of our pipleline. We report accuracy (Acc.), mean recall (MeanR) and MCC on our validation and evaluation data. For a discussion of the results see Sec.~\ref{sec:evaluation:quantitative_results}.}
\centering
\scriptsize
\begin{tabular}{lccc|ccc|ccc}
\toprule
 & \multicolumn{3}{c}{\textbf{Validation Data I}} & \multicolumn{3}{c}{\textbf{Validation Data II}} & \multicolumn{3}{c}{\textbf{Evaluation Data}} \\
\cmidrule(lr){2-4} \cmidrule(lr){5-7} \cmidrule(lr){8-10}
\textbf{Data} & \textbf{Acc.} & \textbf{MeanR} & \textbf{MCC} & \textbf{Acc.} & \textbf{MeanR} & \textbf{MCC} & \textbf{Acc.} & \textbf{MeanR} & \textbf{MCC} \\
\midrule
Binary 3x3 Matrices (30 Epochs) & 96.4 & 93.9 & 88.4 & 95.2 & 92.8 & 85.1 & 91.3 & 88.1 & 75.9 \\
Label GCN (10 Epochs) & 98.0 & 97.9 & 97.0 & 97.9 & 98.1 & 96.9 & 97.2 & 96.3 & 95.6 \\
Permutation GCN (10 Epochs) & 97.8 & 97.2 & 96.8 & 98.7 & 98.7 & 98.2 & 95.3 & 92.7 & 93.3 \\
Sequences GCN (20 Epochs)  &94.0 & 91.5 & 92.2 & 96.9 & 95.0 & 95.6 & 82.6 & 77.3 & 75.9 \\
\bottomrule
\end{tabular}
\label{tab:tab2}
\end{table*}

\begin{figure*}[!ht]
    \centering
    \includegraphics[width=0.91\textwidth]{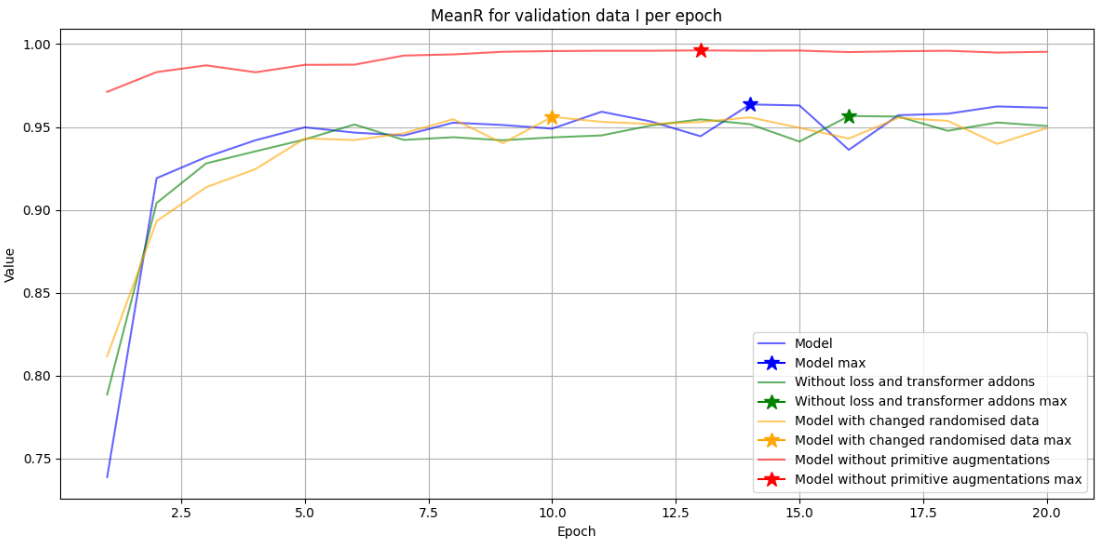}
    \hfill
    \includegraphics[width=0.91\textwidth]{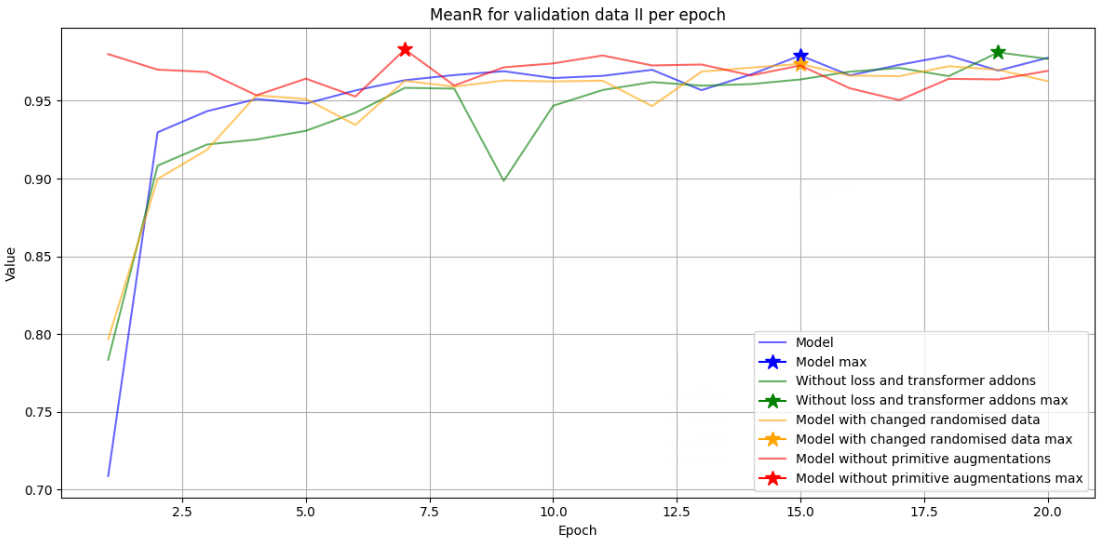}
    \hfill
    \includegraphics[width=0.91\textwidth]{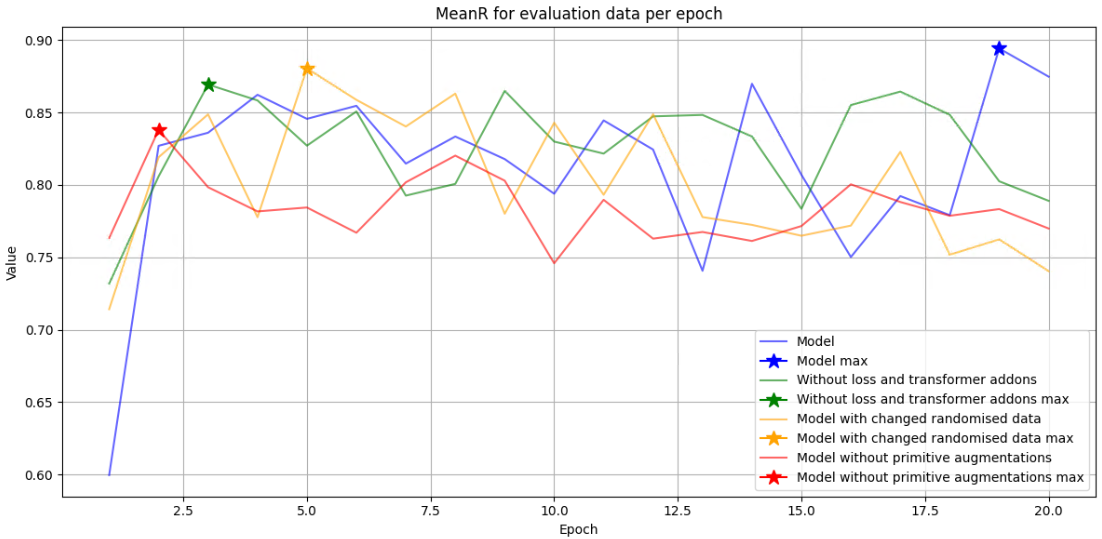}
    \caption{Quantitative results: Visualization of the mean recall (MeanR) on Validation I, Validation II and Evaluation. For further quantitative results see Tab.~\ref{tab:tab2}.}
    \label{fig:3}
\end{figure*}

\subsection{Graph dataset and training strategy}

The training dataset includes three vehicle types: a train, an airplane and a boat. The evaluation dataset contains a previously unseen vehicle—a car (see Fig.~\ref{fig:1}a). Each vehicle consists of specific components (casing, top, bottom, front, screw, wheel) forming a graph structure that encodes part relationships. In total, four distinct graphs are used. The graph size \( n\) denotes the number of components in the largest graph within the dataset, i.e. the graph with the most components. The largest graph in the case study of the toy dataset, corresponding to the train, contains $n=13$ components (plane=8, boat=9, car=13). This corresponds to the size of the predicted adjacency matrix.

The ground truth includes the graph structure as a binary adjacency matrix and GCN output labels for component relations, permutations and assembly sequence, all as $13 \times 13$ matrices.
We define three labels for the component relations (to join, to insert and to screw), four permutation types, such as single connection only (e.g., for screws) or multiple connections possible (e.g., for housing parts). We employ four sequence labels to represent four consecutive assembly steps. Attachment parts (top, bottom, front) are encoded with binary masks and optimized via group loss.

The transformer is trained from images of individual components which have been pre-extracted
using the above Mask R-CNN pipeline. This way, we generate approximately $50$ distinct
instances of real-word image data available per component and vehicle.
This is a very limited amount of data, which can easily lead to
overfitting of a model as powerful as a transformer, as it can just memorize specific configurations~\cite{rong2020dropedge,fang2023dropmessage}.
To address this, we augment the component input images with synthetically
rendered input data we call primitives,
while keeping the structural layout of the target output graph the same, see Fig.\ref{fig:1}d.
For symmetric housing parts (Fig.\ref{fig:1}c – Part A), COCO shape templates \cite{lin2014microsoft} are used to render the inputs.
In this way, we add $400$ additional input images per component to the training data, and one more is generated with
completely abstract shapes (circles/rectangles, see Fig.\ref{fig:1}d, bottom).

For training, we then draw a total of $675,000$ random sets of inputs per graph from both original and augmented sources.
The training strategy also includes additional augmentation techniques: graph sizes and node layouts are randomly altered, adversarial perturbations are introduced and structural graph augmentations are applied.
Every fourth graph is randomly downsampled to a smaller size to increase variance. To ensure diversity, we oversample so that both abstract shapes and real-world images each constitute 5\% of the training dataset. 

\subsection{Implementation details}
We use a pretrained ResNet-50 backbone for both Faster R-CNN and Mask R-CNN. Faster R-CNN is fine-tuned to detect six component classes: casing, wheel, screw, front, bottom and top. Training runs for 20 epochs using SGD (learning rate 0.005, momentum 0.9, weight decay 0.0005). Visual features are extracted using the CLIP ViT-B/32 model; its 512-dimensional output is linearly projected to 36 dimensions. 
We apply AdamW (learning rate 0.001, weight decay 0.001) and 0.1 dropout.
To weight the different loss terms, we use $1.0$ for the focal loss, $10^{-2}$ for the group loss,
and $2.5 \cdot 10^{-4}$ for the Laplacian regularization loss.
A compact transformer (2 layers, 2 heads, hidden size 72) processes the to 36DIM linearly reduced CLIP features. An MLP maps sequences of length 3 to a 3×3 adjacency matrix using batch size 50. During postprocessing, we use a confidence threshold of \(\mathcal{P} \approx 98\%\) to reconstruct the full adjacency matrix from the predicted subgraphs. The Siamese network is trained separately using Adam (learning rate 0.001). The full model runs on a single Nvidia RTX 3060 GPU.

\subsection{Evaluation metrics}
Several scene graph generation tasks exist. In predicate classification (PredCLS) object labels and bounding boxes are given; only relationships must be predicted. Scene graph classification (SGCLS) adds object label prediction, while scene graph detection (SGDET) also requires bounding box prediction. Our task is closest to PredCLS—focusing on relationships—but differs from all three, as bounding boxes are also inferred \cite{cong2023reltr}.

\begin{figure*}[b]%[!ht]%[t]
    \captionsetup{font=footnotesize}
    \centering
    \includegraphics[width=0.8\textwidth]{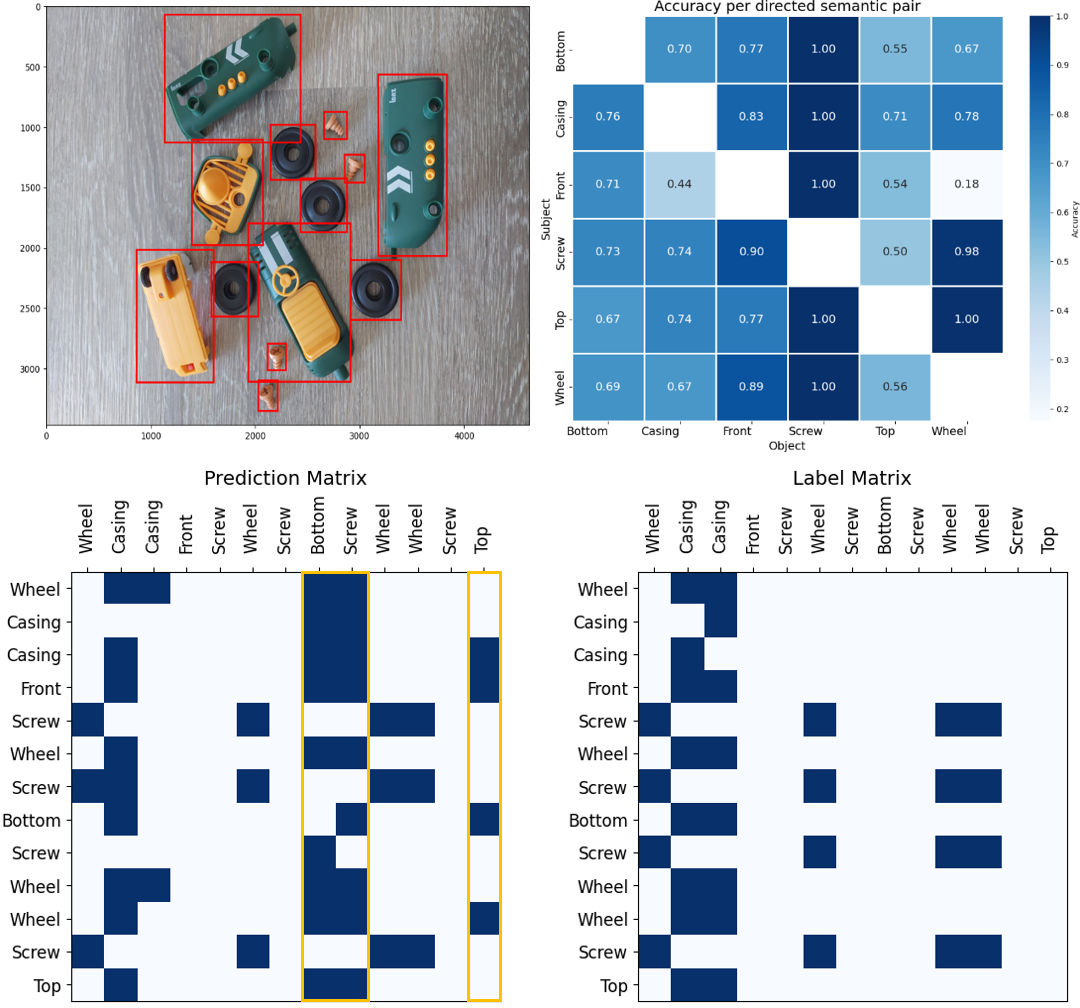}
    \caption{Qualitative results - The unknown vehicle is shown at the top left, followed at the bottom by its binary ground truth (right) and the corresponding detections (left). The two orange rectangles indicate failure cases as discussed in~\cref{sec:evaluation:qualitative_results}. Additionally, the confusion matrix of the evaluation dataset is presented to illustrate overall performance at the top right.}
    \label{fig:4}
\end{figure*}

Assuming accurate detection via Faster R-CNN, we evaluate only the predicted graph structure by comparing predicted and ground truth adjacency matrices. To address class imbalance, we report accuracy, mean recall and the Matthews Correlation Coefficient (MCC) \cite{Tang2020UnbiasedSGG,chicco2021evaluating}:

{\small
\begin{equation}
\text{MCC} = \frac{tp \times tn - fp \times fn}{\sqrt{(tp + fp)(tp + fn)(tn + fp)(tn + fn)}}.
\end{equation}
}

\subsection{Quantitative results and ablations}
\label{sec:evaluation:quantitative_results}

Due to the custom dataset, direct comparison with standard scene graph methods such as Motifs, Graph R-CNN, or RelTR is not feasible, as these require fully triplet-labeled datasets \cite{yang2018graph,cong2023reltr,zellers2018neural}. Instead, we assess our architecture via ablation studies. To analyze overfitting, we use two validation sets: one matching the training set and another with the same vehicles but different images than used during training. A fourth, unseen vehicle serves as the evaluation set.

The model performs well on both validation sets and yields reasonable results on the unseen vehicle despite limited data (see Fig.~\ref{fig:4} and Sec.~\ref{sec:evaluation:qualitative_results}). Geometric shape representations help align unseen samples with the training distribution. Without them recognition of the unknown vehicle fails.

Fig.~\ref{fig:3} shows training dynamics: Overfitting occurs after a few epochs, as indicated by the declining red and yellow curves and the fluctuating blue and green curves in the bottom subfigure of Fig.~\ref{fig:3}. Primitive augmentations enhance robustness. On the evaluation data the green curve (no loss regularization, EGT or depthwise convolutions) performs worse than the blue curve with all enhancements. But their influence is inherently limited, as only few parameters are added relative to the core Transformer and focal loss architecture. Performance is also sensitive to the data sampled graph constellations as shown by the divergence between the similarly configured yellow and blue curves (see bottom sub-figure of Fig.~\ref{fig:3}).

We separately train models for adjacency, label, permutation and sequence prediction to better focus on each task. As shown in Table~\ref{tab:tab2} label classification performs best due to fewer classes. Permutation and sequence prediction each use four classes and are more challenging. Sequence prediction is particularly challenging, as individual pairs lack sufficient information, requiring GCN-based message passing to capture relational context. As shown in Table~\ref{tab:tab2}, it yields the lowest accuracy and strongly depends on the quality of the adjacency matrices. Due to weak performance, the evaluation focuses on the individual models and following only in the binary prediction model.

\subsection{Qualitative results}
\label{sec:evaluation:qualitative_results}

As shown in Fig.~\ref{fig:4}, it is notable that—even with only three training vehicles—the model generates approximate predictions for an unseen vehicle and differentiates well between subjects and objects. Transformers typically require large datasets and identifying familiar components like wheels within unfamiliar graph structures is difficult due to strong contextual bias. This study can be seen as conceptual exploration of improving graph understanding in data-scarce settings. Rather than achieving strong results by overfitting to known vehicle types, we prioritize approaches that target generalization. Fig.~\ref{fig:4} shows a binary prediction of a directed graph. For instance as illustrated by the orange rectangles in Fig.~\ref{fig:4}, a screw is misidentified, and one casing is mistaken for both top and front parts, causing cascading errors. The confusion matrix shows, that attachment and housing components are the hardest to classify correctly.

\section{\uppercase{Conclusion}}

In this paper, we presented an approach for generating scene graphs from a very small dataset without relying on semantic prior knowledge. This is crucial in component assembly, where many parts may belong to unseen categories and labeled data is not given. Notably, our method is capable of approximating relationships even for components outside the training distribution. However, the proposed approach has certain limitations. The reliance on a transformer-based architecture introduces increased model complexity and higher computational costs; its scalability is more suitable for assembly tasks with larger numbers of components. Moreover, the separation between structural prediction and relational characterization can improve modularity and the accuracy of binary predictions of part membership, yet it can also lead to error propagation, as inaccuracies in the predicted adjacency matrix negatively affect subsequent predictions of labels, permutations, and sequences. Finally, despite the use of geometric cues for reasoning without semantic priors, ambiguous or visually similar components remain challenging, particularly in sparse or noisy input scenarios.

Even with limited performance, these cases provide useful insights. But using datasets for data-driven approaches that are too small is not recommendable and the model would clearly benefit from a larger dataset, potentially enabling direct multiclass prediction. With sufficient data, set-based triplet prediction could be explored and compared to standard scene graph methods. Recurrent GCNs and Siamese CNNs show potential for learning part relations and improving generalization on small datasets, especially when adjacency can be learned directly from the model. Finally, this work forms a basis for 3D scene graph generation, with 3D data offering richer features to reduce input ambiguity and enhance prediction accuracy.

\section*{\uppercase{Acknowledgements}}
U.W. and B.G. acknowledge funding by the Deutsche Forschungsgemeinschaft (DFG, German Research Foundation) under Germany's Excellence Strategy – EXC 2117 – 422037984.

\bibliographystyle{apalike}
{\small
\bibliography{references}}

\end{document}